\pdfoutput=1

\documentclass[11pt,table,svgnames]{article}

\usepackage[preprint]{acl}

\usepackage{times}

\usepackage{booktabs}
\usepackage{latexsym}
\usepackage{tikz}
\usepackage{arydshln}

\usepackage[T1]{fontenc}

\usepackage[utf8]{inputenc}

\usepackage{microtype}

\usepackage{inconsolata}

\usepackage{graphicx}
\usepackage{multirow}

\title{Unleashing the Reasoning Potential of Pre-trained LLMs by \\ Critique Fine-Tuning on One Problem}

\author{
Yubo Wang\textsuperscript{1,2}, Ping Nie\textsuperscript{5}, Kai Zou\textsuperscript{3}, Lijun Wu\textsuperscript{4}, Wenhu Chen\textsuperscript{1,2} \\
\textsuperscript{1}University of Waterloo, 
\textsuperscript{2}Vector Institute,
\textsuperscript{3}Netmind.AI, 
\textsuperscript{4}Shanghai AI Lab,
\textsuperscript{5}Independent\\
\vspace{2ex}
\url{https://tiger-ai-lab.github.io/One-Shot-CFT}
}

\begin{document}
\maketitle

\begin{abstract}
We have witnessed that strong LLMs like Qwen-Math, MiMo, and Phi-4 possess immense reasoning potential inherited from the pre-training stage. With reinforcement learning (RL), these models can improve dramatically on reasoning tasks. Recent studies have shown that even RL on a single problem~\cite{one-shot-rlvr} can unleash these models' reasoning capabilities. However, RL is not only expensive but also unstable. Even one-shot RL requires hundreds of GPU hours. This raises a critical question: Is there a more efficient way to unleash the reasoning potential of these powerful base LLMs?
In this work, we demonstrate that Critique Fine-Tuning (CFT) on only one problem can effectively unleash the reasoning potential of LLMs. Our method constructs critique data by collecting diverse model-generated solutions to a single problem and using teacher LLMs to provide detailed critiques. We fine-tune Qwen and Llama family models, ranging from 1.5B to 14B parameters, on the CFT data and observe significant performance gains across diverse reasoning tasks. For example, with just 5 GPU hours of training, Qwen-Math-7B-CFT show an average improvement of 15\% on six math benchmarks and 16\% on three logic reasoning benchmarks. These results are comparable to or even surpass the results from RL with 20x less compute.
Ablation studies reveal the robustness of one-shot CFT across different prompt problems. These results highlight one-shot CFT as a simple, general, and compute-efficient approach to unleashing the reasoning capabilities of modern LLMs.

\end{abstract}

\begin{figure*}[!t]
    \centering
    \includegraphics[width=0.95\textwidth]{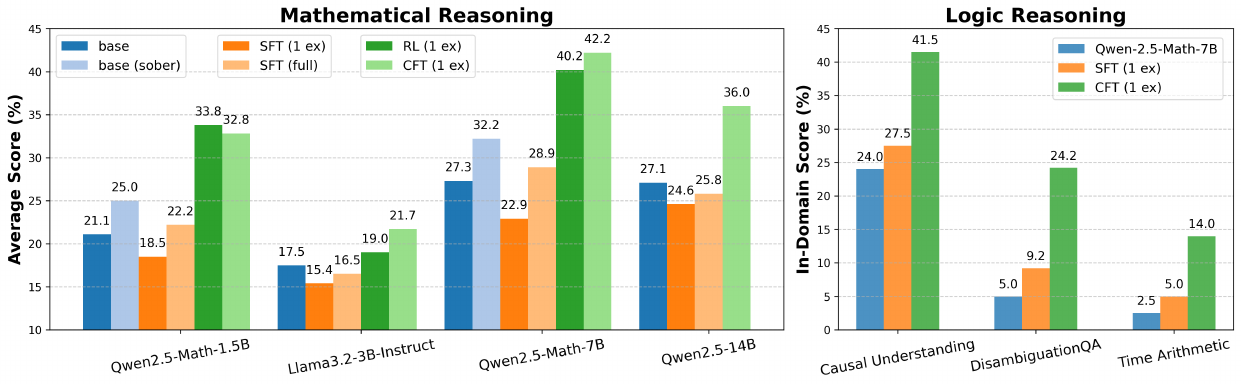}
    \vspace{-1ex}
    \caption{
        \textbf{One-shot CFT consistently improves mathematical and logical reasoning.}
\textbf{Left:} Average accuracy (\%) on six mathematical reasoning benchmarks for Qwen and Llama models, comparing base, SFT, RLVR, and CFT with only one training example.
\textbf{Right:} In-domain accuracy (\%) on three logic reasoning benchmarks (BBEH subtasks) for Qwen2.5-Math-7B.
Across both domains, CFT with a single problem significantly outperforms standard supervised fine-tuning and matches or exceeds reinforcement learning with much lower compute.
    }
    \label{fig:teaser}
\end{figure*}

\section{Introduction}
Large language models (LLMs) have recently achieved impressive results on mathematical and scientific reasoning tasks~\cite{gpt4,qwen3,math,minerva,mmlupro,supergpqa}, demonstrating strong generalization and reasoning capabilities. 
Among various post-training methods, reinforcement learning with verifiable rewards (RLVR)~\citep{deepseek-r1} has shown particular promise in enhancing reasoning ability by enabling models to learn through trial-and-error exploration~\cite{simplerl-zoo,general-reasoner}. Interestingly, recent studies reveal that even a single training example can significantly improve model performance through 1-shot RLVR~\cite{one-shot-rlvr}. These findings suggest that base models inherently possess substantial reasoning potential, which can be effectively unleashed with minimal and targeted training signals.

However, RLVR methods suffer from two major drawbacks. First, it is highly resource-intensive, often requiring over 100 GPU hours to fine-tune even a 7B model on a single problem~\cite{one-shot-rlvr}. Second, due to issues like transient non-stationarity and plasticity loss, RLVR often exhibits unstable or unreliable training dynamics~\cite{dang2025reinforcement,goldie2024can,igl2020transient}. 

Another popular post-training method is supervised fine-tuning (SFT). While SFT is more stable than RLVR, it typically depends on large volumes of high-quality data to prevent overfitting. For instance, Phi-4-Reasoning~\cite{phi4-reasoning} was fine-tuned on 16B tokens (over a million examples), and AceMath~\cite{acemath} used 1.6M curated examples. However, in many reasoning tasks beyond mathematics, large-scale, high-quality datasets are scarce, which can hinder the effectiveness of SFT and increase the risk of overfitting on small datasets.

Recently, Critique Fine-Tuning (CFT) has emerged as a promising alternative~\cite{cft}. By enabling models to learn from critiques of diverse incorrect solutions, CFT can enhance the model's exposure to varied reasoning patterns and mitigates overfitting. Specifically, CFT introduces diversity by allowing teacher models to critique a wide range of candidate answers to a given problem. This exposes the LLM to multiple perspectives and error types, thereby more effectively unleashing its reasoning potential. This leads to the question: 

\emph{Can critiques from a single problem suffice to unleash LLMs’ reasoning potential, achieving RLVR-level effectiveness with minimum cost?}

\begin{figure*}[t]
    \centering
    \includegraphics[width=\linewidth]{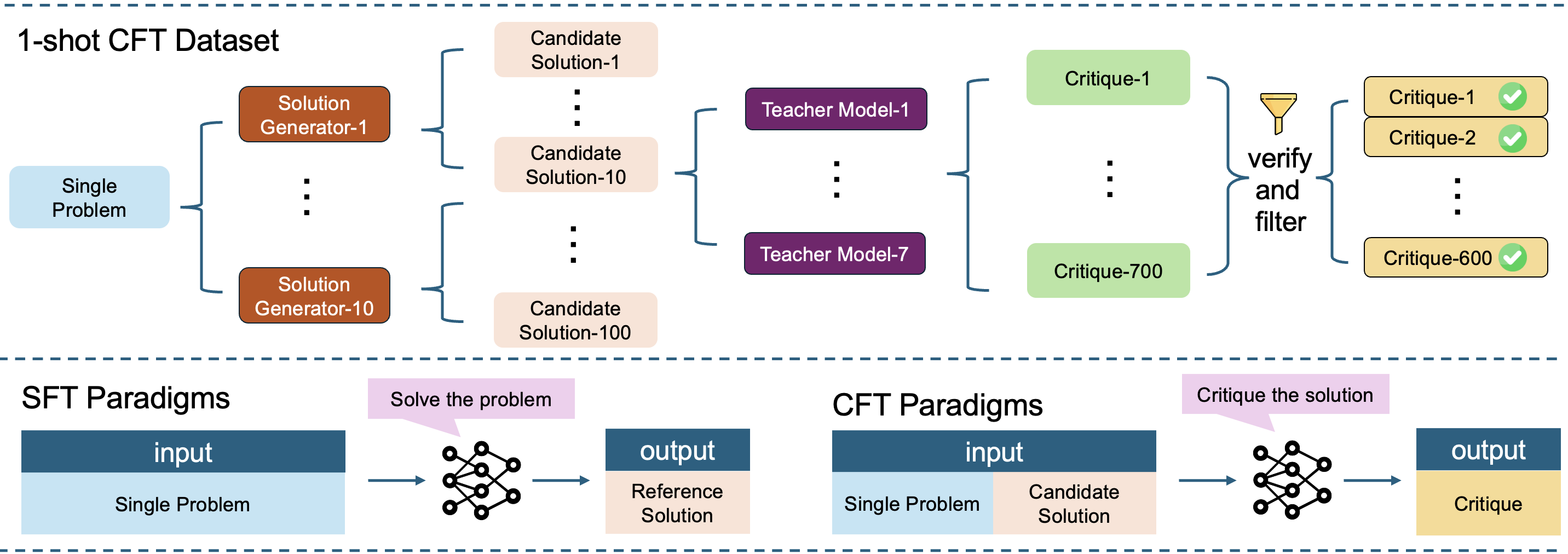}
    \vspace{-2ex}
    \caption{
Overview of the 1-shot CFT dataset construction and the key difference between SFT and CFT training. \textbf{Top:} Candidate solutions to a single math problem are generated, critiqued, and filtered to form the training set. \textbf{Bottom:} Comparison of training paradigms: \textbf{(left)} SFT supervises the model to generate the reference solution; \textbf{(right)} CFT trains the model to critique a candidate solution, encouraging deeper reasoning and error analysis.
}
    \label{fig:intro}
\end{figure*}

In this work, we systematically investigate one-shot CFT as a general and compute-efficient post-training approach for both mathematical and logical reasoning tasks. As shown in Figure~\ref{fig:intro}, we first generate diverse candidate solutions to a single problem using multiple open-source models, then employ strong teacher LLMs to provide detailed critiques for each solution. These high-quality critiques are filtered and used to fine-tune Qwen and Llama models (1.5B--14B parameters) on compact yet information-rich datasets.

Our experiments demonstrate that one-shot CFT yields substantial performance gains on both fronts. On six standard math reasoning benchmarks, our approach achieves up to a 15\% absolute improvement in average accuracy (e.g., boosting Qwen2.5-Math-7B from 27\% to 42\%) with only 5 GPU hours. Notably, on Minerva~\citep{minerva}, OlympiadBench~\citep{olympiadbench} and AMC-23, one-shot CFT can improve the accuracy by more than 20\%. On logic reasoning, we observe a 16\% average accuracy gain across three representative subtasks from BIG-Bench Extra Hard (BBEH)~\citep{bbeh}, including \textit{Causal Understanding}, \textit{DisambiguationQA}, and \textit{Time Arithmetic}. Further ablation studies confirm the robustness of one-shot CFT across different seed problems and model combinations.

Overall, our findings highlight one-shot CFT as a simple, robust, and compute-efficient paradigm for unleashing the reasoning capabilities of modern LLMs in both mathematical and logical domains.

\section{Related Work}

\subsection{Advances in Post-Training for Reasoning}
Post-training~\citep{instructgpt} is a crucial step in aligning pre-trained language models (LLMs) to solve specific tasks. Recently, there has been a surge of interest in post-training methods for improving reasoning performance, particularly in mathematical and coding problems~\citep{toshniwal2024openmathinstruct,mammoth,deepseekmath,deepseek-r1}. These methods have demonstrated significant improvements, with approaches like DeepSeekMath~\citep{deepseekmath} achieving strong performance on benchmarks such as MATH-500~\citep{math} and AMC~\citep{minerva}.

To make post-training more efficient, researchers have explored methods that require minimal data. S1~\citep{s1} and LIMO~\citep{limo} showed notable advances by leveraging 1000 training examples to enhance reasoning capabilities. Building on this, 1-shot RLVR~\citep{one-shot-rlvr} reduced data requirements to a single example, demonstrating that base LLMs possess latent reasoning abilities that can be efficiently activated with minimal supervision. These methods highlight the potential of leveraging small amounts of data to unlock reasoning capabilities in LLMs.

Our work builds on this paradigm of efficient post-training but introduces critique fine-tuning (CFT)~\citep{cft} as an alternative to supervised fine-tuning (SFT) and reinforcement learning (RL). Unlike SFT, which risks overfitting to reference solutions, CFT encourages models to analyze errors and reason critically by exposing them to diverse perspectives and error types. This approach has been shown to generalize better, especially in data-scarce scenarios.

\subsection{Challenges of RLVR}
Recent studies have raised concerns about the robustness and reproducibility of RLVR-based methods. \citet{sober} emphasized that many reported RLVR improvements rely on unstable evaluation practices, such as single-seed evaluations or small benchmark sizes, which can lead to misleading conclusions about reasoning progress. They also highlighted the importance of reproducible evaluation frameworks to ensure fair comparisons across methods.

Similarly, \citet{rethink-rlvr} demonstrated that RLVR gains often depend on spurious reward signals or model-specific priors rather than genuine reasoning improvements. For example, random or incorrect rewards were shown to produce significant performance boosts in Qwen2.5-Math models, but these effects failed to generalize to other model families like Llama or OLMo. This suggests that RLVR often amplifies pre-existing capabilities rather than teaching new reasoning skills, raising questions about its generalizability and scalability.

In this context, CFT provides an alternative approach by leveraging critiques as training signals. It shows strong performance not only on mathematical reasoning tasks like MATH-500~\citep{math} and Minerva~\citep{minerva}, but also on logic reasoning tasks from the BIG-Bench Extra Hard (BBEH) benchmark~\citep{bbeh}, demonstrating potential for generalizing across diverse reasoning domains.

\section{Method}

In this section, we will detail our dataset construction and training scheme.
\subsection{One-shot CFT Dataset Construction}
To systematically assess one-shot CFT, we construct a suite of critique datasets derived from a single seed problem, following the one-shot RLVR research protocol. Our process is as follows:

\noindent\textbf{Seed Problem Selection.}
We select seed math problems from the DeepScaleR subset, following the setting of previous one-shot RLVR studies. For ease of comparison, we focus on four representative problems, $\pi_1$, $\pi_2$, $\pi_{13}$, and $\pi_{1209}$, which were also analyzed in prior work. The full content of these seed problems is provided in Appendix~\ref{app:seed-problems}.

\noindent\textbf{Candidate Solutions Generation.}
For each seed problem, we use 10 open-source models as solution generators, including Qwen2.5-Math-7B-Instruct~\cite{qwen2-math}, Qwen3-4B/8B/14B/32B~\cite{qwen3}, MiMo-7B-SFT~\cite{mimo}, MiMo-7B-RL, DeepSeek-R1-Distill-Qwen-32B~\cite{deepseek-r1}, Phi-4-reasoning~\cite{phi4-reasoning}, and Phi-4-reasoning-plus. Each generator provides 10 solutions for the seed problem, resulting in 100 various candidate solutions. (see Fig.~\ref{fig:teaser})

\noindent\textbf{Teacher Critique Annotation.}
We then solicit critiques for each candidate solution from 7 high-performing, proprietary teacher models: Claude-3-7-Sonnet~\cite{claude37sonnet}, Claude-3-5-Sonnet, GPT-4.1-Mini~\cite{gpt41}, GPT-4.1, GPT-4o~\cite{gpt4}, O3-Mini~\cite{openaio3}, and O1-2024~\citep{o1}.

\begin{figure*}[ht]
    \centering
    \includegraphics[width=0.98\textwidth]{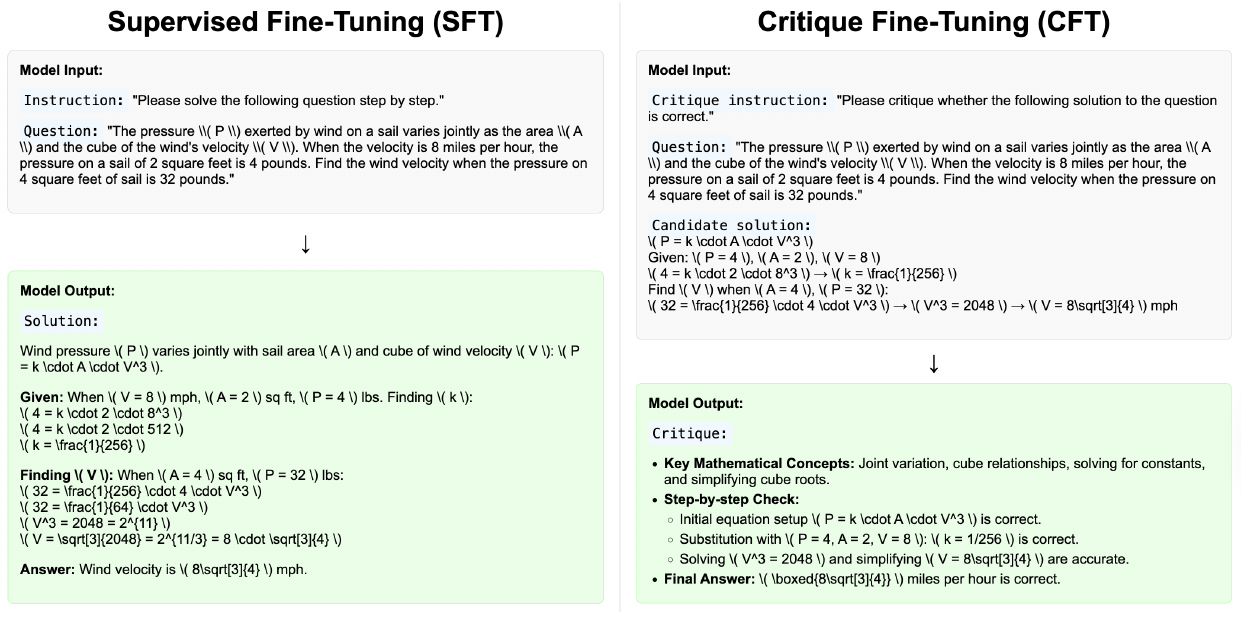}
    \caption{Comparison between Supervised Fine-Tuning (SFT) and Critique Fine-Tuning (CFT). SFT generates solutions directly, while CFT critiques candidate solutions for correctness.}
    \label{fig:sft_cft_compare}
\end{figure*}

\subsection{Dataset Statistics}

For each seed problem, we start with 700 critiques (100 candidate solutions $\times$ 7 teacher critiques). 
After filtering out incorrect or inconsistent critiques, we remove 43, 16, 22, and 68 examples for $\pi_1$, $\pi_2$, $\pi_{13}$, and $\pi_{1209}$, respectively. To ensure fair comparison across different seeds, we further remove the longest and shortest samples by length and subsample the remaining data to construct a unified training set of 600 critiques per seed problem.
Table~\ref{tab:data_statistics} summarizes the dataset statistics, including average input and output token counts and difficulty ratings. The problems vary in difficulty: $\pi_2$ and $\pi_{13}$ are easy, $\pi_1$ is medium, and $\pi_{1209}$ is hard.

\begin{table}[ht]
    \setlength{\tabcolsep}{2pt} 
    \centering
    \small
    \begin{tabular}{p{1.5cm}p{1.5cm}p{1.2cm}p{1.2cm}p{1.2cm}}
        \toprule
        \textbf{Training} & \textbf{Problem} & \textbf{Input} & \textbf{Output} & \textbf{Difficulty}\\ 
        \textbf{Dataset} & \textbf{ID} & \textbf{Tokens} & \textbf{Tokens} & \textbf{Level}\\
        \midrule
        dsr-cft-p0 & $\pi_1$ & 736 & 614 & Medium \\
        dsr-cft-p1 & $\pi_2$ & 779 & 621 & Easy \\
        dsr-cft-p2 & $\pi_{13}$ & 1763 & 1024 & Easy \\
        dsr-cft-p3 & $\pi_{1209}$ & 1136 & 992 & Hard \\
        \bottomrule
    \end{tabular}
    \caption{Statistics of the training datasets for one-shot CFT. Each dataset contains 600 critique examples per seed problem. Token counts are averaged across all examples in the dataset.}
    \label{tab:data_statistics}
\end{table}

\subsection{Training}

Following the Critique Fine-Tuning (CFT)~\citep{cft}, each training instance is constructed by concatenating the original problem and a candidate solution as the model input, with the corresponding teacher-provided critique serving as the target output. Specifically, each sample takes the form $(x, y) \rightarrow c$, where $x$ denotes the seed problem, $y$ is a candidate solution from a student model, and $c$ is the critique provided by a teacher model. During training, the model learns to generate $c$ given the concatenated $(x, y)$ as input. As illustrated in Figure~\ref{fig:sft_cft_compare}, unlike Supervised Fine-Tuning (SFT), which directly trains the model to generate solutions, CFT emphasizes critiquing candidate solutions for correctness and step-by-step reasoning. Detailed templates and training examples are provided in Appendix~\ref{appendix:training_data_format}.

We adopt full-parameter instruction tuning for all experiments. Models are trained with a learning rate of $5 \times 10^{-6}$, using a cosine learning rate schedule and a warmup ratio of $0.1$. The global batch size is set to 512. To ensure fair comparison, all hyperparameters are kept consistent across different model architectures and problem seeds.

\section{Experiments on Math Reasoning}

\begin{table*}[t]
\centering
\small
\resizebox{\textwidth}{!}{
\begin{tabular}{ll|cccccc|c}
\toprule
\textbf{Model} & \textbf{Method} & \textbf{Math-500} & \textbf{Minerva} & \textbf{Olympiad} & \textbf{AIME24} & \textbf{AIME25} & \textbf{AMC23} & \textbf{AVG} \\
\midrule
\multirow{6}{*}{Qwen2.5-Math-1.5B}
  & base      & 35.8 & 11.0 & 22.1 & 15.0 & 2.5 & 40.0 & 21.1 \\
  & base (sober)  &  51.7 & 11.3  & 26.0  &  11.3  &  5.7  & 44.0 & 25.0  \\
  & SFT (1 ex)    & 37.2 & 9.6  & 22.7 & 3.1  & 0.0 & 38.3 & 18.5 \\
  & SFT (full)    & 49 & 14.3 & 23.2 & 7.9  & 2.1 & 35.8 & 22.2 \\
  & RL (1 ex)     & 72.4 & 26.8 & 33.3 & 11.7 & 7.1 & 51.6 & \textbf{33.8} \\
  & \textbf{CFT (1 ex)} & 66.6 & 30.1 & 30.4 & 10.4 & 8.8 & 50.6 & 32.8 \\
\rowcolor{LightCyan}  & $\Delta$ = CFT - base & +30.8 & +19.1 & +8.3 & -4.6 & +6.3 & +10.6 & +11.7 \\
\midrule
\multirow{6}{*}{Llama3.2-3B-Instruct}
  & base      & 40.8 & 15.8 & 13.2 & 8.3  & 1.7 & 25.3 & 17.5 \\
  & SFT (1 ex)    & 41.4 & 13.2 & 11.7 & 2.7  & 0.0 & 23.2 & 15.4 \\
  & SFT (full)    & 43.2 & 14.7 & 12.1 & 3.1  & 1.7 & 24.3 & 16.5 \\
  & RL (1 ex)     & 45.8 & 16.5 & 17.0 & 7.9  & 1.2 & 25.3 & 19.0 \\
  & \textbf{CFT (1 ex)} & 49.0 & 21.0 & 15.3 & 9.2  & 2.9 & 32.5 & \textbf{21.7} \\
\rowcolor{LightCyan}  & $\Delta$ = CFT - base & +8.2 & +5.2 & +2.1 & +0.9 & +1.2 & +7.2 & +4.2 \\
\midrule
\multirow{6}{*}{Qwen2.5-Math-7B}
  & base      & 58.6 & 17.3 & 17.5 & 16.7 & 10.8 & 43.1 & 27.3 \\
  & base (sober)   & 64.3 &  17.3 &  26.0 & 20.7 & 8.7  & 56.2 & 32.2 \\
  & SFT (1 ex)    & 53.8 & 14.3 & 18.2 & 12.1 & 6.7  & 32.5 & 22.9 \\
  & SFT (full)    & 58.6 & 24.6 & 27.6 & 10.0 & 7.1  & 45.3 & 28.9 \\
  & RL (1 ex)     & 79.2 & 27.9 & 39.1 & 23.8 & 10.8 & 60.3 & 40.2 \\
  & \textbf{CFT (1 ex)} & 76.4 & 40.4 & 39.3 & 18.8 & 14.6 & 63.4 & \textbf{42.2} \\
\rowcolor{LightCyan}  & $\Delta$ = CFT - base & +17.8 & +23.1 & +21.8 & +2.1 & +3.8 & +20.3 & +14.9 \\
\midrule
\multirow{5}{*}{Qwen2.5-14B}
  & base      & 60.4 & 22.4 & 27.9 & 3.8  & 3.8 & 44.1 & 27.1 \\
  & SFT (1 ex)    & 63.8 & 19.5 & 20.9 & 5.0  & 1.2 & 36.9 & 24.6 \\
  & SFT (full)    & 65.2 & 24.2 & 22.7 & 2.6  & 1.7 & 38.3 & 25.8 \\
  & \textbf{CFT (1 ex)} & 71.2 & 43.8 & 34.8 & 12.5 & 8.3 & 45.3 & \textbf{36.0} \\
\rowcolor{LightCyan}  & $\Delta$ = CFT - base & +10.8 & +21.4 & +6.9 & +8.7 & +4.5 & +1.2 & +8.9 \\
\bottomrule
\end{tabular}
}
\caption{Performance (\%) on mathematical benchmarks. The base results are measured using the same prompt and evaluation setting with SFT and CFT. The base (sober) is taken from~\citet{sober} with a more comprehensive evaluation. The RL (1 ex) results are from \citet{one-shot-rlvr}. The delta rows show the performance difference between CFT (1 ex) and the base.}
\label{tab:main}
\end{table*}

\subsection{Setup}
We conduct our experiments on four backbone models: Qwen2.5-Math-1.5B, Qwen2.5-Math-7B, Llama-3.2-3B-Instruct, and Qwen2.5-14B. For seed question selection, we follow the protocol established in one-shot RLVR studies and choose the same four representative problems: $\pi_1$, $\pi_2$, $\pi_{13}$, and $\pi_{1209}$. The corresponding CFT training datasets are denoted as \texttt{dsr-cft-p0}, \texttt{dsr-cft-p1}, \texttt{dsr-cft-p2}, and \texttt{dsr-cft-p3}.

To facilitate a fair comparison with supervised fine-tuning (SFT) methods, we employ the full DeepScaleR dataset (40.9K examples) as the training data for our Full SFT baseline. Additionally, for the one-example SFT (SFT-1ex) condition, we select $\pi_1$ as the seed problem and use the same 7 closed-source API models to generate 100 diverse solutions. We then verify all 700 generated solutions against the ground-truth answer, retaining 600 correct responses for our final SFT (1 ex) dataset.

For evaluation, we include sober baseline scores provided by \citet{sober}, which reassess prior backbone results under a standardized framework. These baselines highlight underestimation in earlier evaluations of Qwen models, providing a more rigorous reference point for comparison. We evaluate all models on six standard mathematical reasoning benchmarks: MATH-500, Minerva Math, OlympiadBench, AIME25, AIME24, and AMC23. To ensure statistical stability for the smaller benchmarks (AIME25, AIME24, AMC23), we repeat each evaluation 32 times and report the average result as the final score.

\begin{table*}[!t]
\centering
\small
\setlength{\tabcolsep}{3pt}
\begin{tabular}{l|c|cccccc|c}
\toprule
\textbf{Training Data} & \textbf{Seed Score (/100)} & \textbf{Math-500} & \textbf{Minerva Math} & \textbf{Olympiad} & \textbf{AIME25} & \textbf{AIME24} & \textbf{AMC23} & \textbf{AVG} \\
\midrule
baseline & - & 52.6 & 17.3 & 17.5 & 10.8 & 16.7 & 43.1 & 26.3 \\
\midrule
dsr-cft-p0 & 49.0 & \textbf{77.0} & \textbf{40.4} & \textbf{39.3} & 14.6 & 18.8 & 63.4 & \textbf{42.2} \\
dsr-cft-p1 & 93.0 & 72.4 & 35.7 & 32.1 & \textbf{15.8} & \textbf{20.0} & 51.6 & 37.9 \\
dsr-cft-p2 & 83.0 & 77.0 & 33.1 & 39.1 & 12.1 & 13.8 & 57.2 & 38.7 \\
dsr-cft-p3 & 10.0 & 72.6 & 32.4 & 35.4 & 7.1 & 10.4 & 59.7 & 36.3 \\
dsr-cft-p0,p1,p2,p3 & 58.8 & 74.6 & 34.6 & 35.4 & 13.3 & 17.1 & \textbf{65.3} & 40.1 \\
\bottomrule
\end{tabular}
\caption{Comparison of performance (\%) with different seed math problems on Qwen-2.5-Math-7B}
\label{tab:1shot}
\end{table*}


\begin{table*}[t]
\centering
\small
\begin{tabular}{l|cccccc|c}
\toprule
\textbf{Solution Generators}  & \textbf{Math-500} & \textbf{Minerva} & \textbf{Olympiad} & \textbf{AIME25} & \textbf{AIME24} & \textbf{AMC23} & \textbf{Avg}\\
\midrule
1 generator (Phi-4)  & 75.8 & 32.0 & 35.4 & 7.1 & 16.7 & 58.8 & 37.6\\
1 generator (Qwen2.5)  & 74.4 & 30.5 & 35.6 & 9.6 & 17.1 & 64.7 & 38.7\\
10 generators (mixed)   & \textbf{76.4} & \textbf{40.4} & \textbf{39.3} & \textbf{14.6} & \textbf{18.8} & \textbf{63.4} & \textbf{42.2}\\
\bottomrule
\end{tabular}
\caption{
Full ablation results on the diversity of solution generators in one-shot CFT.
}
\label{tab:ablation_diversity_main}
\end{table*}

\subsection{Main Results}

Table~\ref{tab:main} presents the main performance comparison across different training methods, including one-shot Critique Fine-Tuning (CFT), supervised fine-tuning (SFT), and one-shot Reinforcement Learning with Verifiable Reward (RLVR). For validation, we randomly select 500 math problems from the MATH dataset (excluding those in the MATH-500 benchmark) to construct the validation set. During training, all models are checkpointed every 10 steps. The checkpoint with the highest validation score is selected for final evaluation.

\noindent\textbf{CFT significantly improves upon the backbone.}  
Across all model scales, one-shot CFT consistently improves reasoning accuracy over the base models. Even when evaluated against the more rigorous sober baseline scores by \citet{sober}, CFT demonstrates substantial gains. For instance, on Qwen2.5-Math-7B, the backbone accuracy is revised to 32.2\%, and one-shot CFT still achieves 42.2\%, delivering a +10.0 point improvement.

\noindent\textbf{CFT outperforms SFT even with full data.}
Under the same one-shot setting, CFT substantially outperforms SFT. For Qwen2.5-Math-7B, one-shot SFT achieves 22.9\%, while one-shot CFT reaches 42.2\%. Notably, one-shot CFT also surpasses SFT trained on the full dataset (25.6\%), highlighting the superior generalization and reasoning gains from the critique supervision signal.

\noindent\textbf{CFT is competitive with or superior to one-shot RLVR.}
CFT demonstrates stronger performance than RLVR across most settings. On Qwen2.5-Math-7B and Llama-3.2-3B-Instruct, one-shot CFT outperforms RLVR by +2.0 and +2.1 points, respectively.
On Qwen2.5-Math-1.5B, CFT is slightly behind RLVR (by 1 point).


\subsection{Training Efficiency Comparison}

\begin{figure}[ht]
    \centering
    \includegraphics[width=\linewidth]{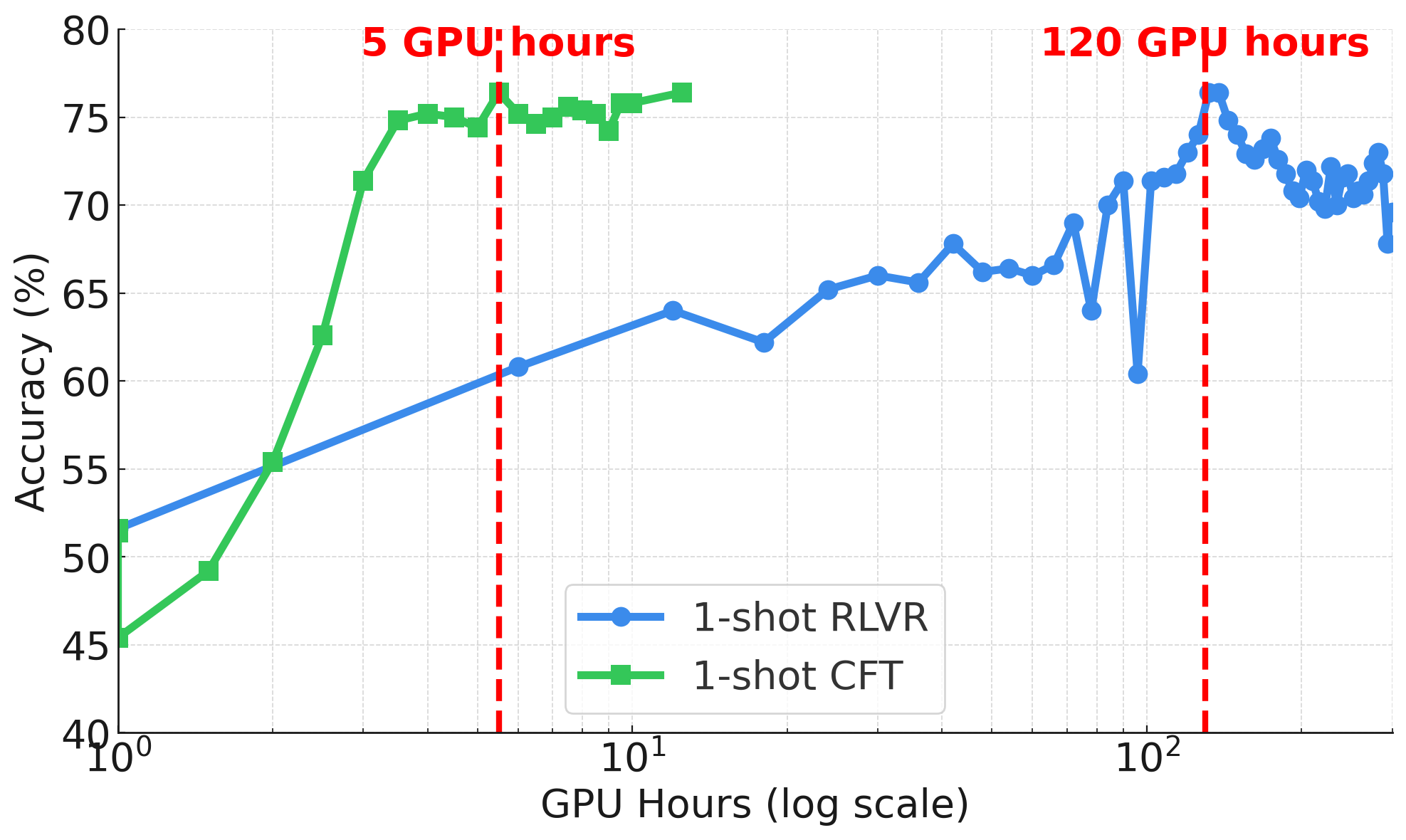}
    \caption{Comparing Model accuracy on Math-500, v.s. the training cost. For the Qwen2.5-Math-7B trained with 1-shot RL and 1-shot CFT.}
    \label{fig:efficient}
    \vspace{-2ex}
\end{figure}

As shown in Figure~\ref{fig:efficient}, one-shot CFT achieves significantly higher training efficiency than one-shot RLVR. With only 5 GPU hours, CFT surpasses 75\% accuracy on the Math-500 and quickly stabilizes. In contrast, RLVR requires over 120 GPU hours to reach a similar level of performance and exhibits greater fluctuations during training.

This efficiency advantage is primarily due to the high computational cost of reinforcement learning, which requires many iterations to propagate reward signals. In contrast, CFT benefits from direct and dense critique supervision, enabling much faster and more stable training. Consequently, one-shot CFT matches or surpasses RLVR performance while using only about 1/15 to 1/20 of the compute.

\subsection{Effectiveness of Seed Examples}

Table~\ref{tab:1shot} compares one-shot CFT performance on datasets from different seed problems. While all seeds are effective, \texttt{dsr-cft-p0} (from seed problem $\pi_1$) achieves the highest average accuracy.

To understand this, we assess the difficulty of each seed by prompting Qwen3-32B to grade 100 candidate solutions from Qwen2.5-Math-7B, using the grading prompt provided in Appendix~\ref{appendix:prompts}. Scores of 1 (correct), 0.5 (partially correct), or 0 (incorrect) are assigned and summed. Seeds of moderate difficulty, such as $\pi_1$, yield a balanced mix of correct and incorrect solutions, enabling richer critiques and more effective learning.

Overall, one-shot CFT is robust to the seed choice, with moderate-difficulty seeds providing the strongest learning signal.

\begin{table*}[ht]
\centering
\small
\begin{tabular}{lccc|c}
\toprule
\textbf{Model} & \textbf{Causal Understanding} & \textbf{DisambiguationQA} & \textbf{Time Arithmetic} & \textbf{Average} \\
\midrule
Qwen2.5-Math-7B & 24.0 & 5.0 & 2.5 & 10.5 \\
\midrule
\multicolumn{5}{c}{\textit{Fine-tuned on 1 problem from Causal Understanding}} \\
SFT (1 ex) & \cellcolor{purple!15}27.5 & 11.7 & 2.0 & 13.7 \\
CFT (1 ex) & \cellcolor{purple!15}\textbf{41.5} & 25.0 & 9.0 & 25.2 \\
\midrule
\multicolumn{5}{c}{\textit{Fine-tuned on 1 problem from DisambiguationQA}} \\
SFT (1 ex) & 20.5 & \cellcolor{purple!15}9.2 & 2.0 & 10.6 \\
CFT (1 ex) & 34.5 & \cellcolor{purple!15}24.2 & 2.5 & 20.4 \\
\midrule
\multicolumn{5}{c}{\textit{Fine-tuned on 1 problem from Time Arithmetic}} \\
SFT (1 ex) & 24.5 & 10.8 & \cellcolor{purple!15}5.0 & 13.4 \\
CFT (1 ex) & 37.0 & \textbf{28.3} & \cellcolor{purple!15}14.0 & 26.4 \\
\midrule
\multicolumn{5}{c}{\textit{Fine-tuned on 3 problems from the above tasks.}} \\
SFT (3 ex) & \cellcolor{purple!15}29.5 & \cellcolor{purple!15}11.7 & \cellcolor{purple!15}6.5 & \cellcolor{purple!15}15.9 \\
CFT (3 ex) & \cellcolor{purple!15}36.5 & \cellcolor{purple!15}\textbf{28.3} & \cellcolor{purple!15}\textbf{15.5} & \cellcolor{purple!15}\textbf{26.8} \\
\bottomrule
\end{tabular}
\caption{
Performance of Qwen2.5-Math-7B on three BIG-Bench Extra Hard (BBEH) logic reasoning subtasks. 
For each subtask, SFT and CFT are performed using a single example from that subtask, and evaluated on all three tasks. 
The in-domain (diagonal) results and all results in the last two rows (merged CFT/SFT) are highlighted. The last two rows show results when merging the three problems into a single three-example training set, evaluating generalization across all three subtasks. Best results in each column are in bold.
}
\label{tab:bbeh-1shot-cft}
\end{table*}

\subsection{Diversity of Candidate Solutions}

To analyze the effect of candidate solution diversity, we compare three settings on the seed problem $\pi_1$. We use a single strong generator (Phi-4-Reasoning-Plus) and a single weaker generator (Qwen2.5-Math-7B-Instruct) to each produce 100 candidate solutions, generate critiques, and perform CFT. Our main method, by contrast, mixes 100 candidate solutions from 10 different generators before collecting critiques and fine-tuning.

As shown in Table~\ref{tab:ablation_diversity_main}, using a single generator yields average benchmark scores of 38.7 (Phi-4-Reasoning-Plus) and 37.6 (Qwen2.5-Math-7B-Instruct). In comparison, our mixed-generator approach achieves a higher average of 42.2. This demonstrates that greater diversity in candidate solutions leads to richer error types and reasoning patterns, enabling more effective critique fine-tuning.

\section{Experiments on Logic Reasoning}
\label{sec:logic-reasoning}

To further evaluate the effectiveness of one-shot CFT beyond mathematical reasoning, we conducted experiments on logic reasoning tasks from the BIG-Bench Extra Hard (BBEH) benchmark. BBEH is a challenging benchmark designed to test advanced reasoning capabilities of large language models. It consists of 23 subtasks, each containing 200 questions, except for the \textit{DisambiguationQA} subtask, which has 120 questions. 

In this section, we present one-shot CFT results on three logic reasoning subtasks: \textit{Causal Understanding}, \textit{DisambiguationQA}, and \textit{Time Arithmetic}. These tasks cover diverse reasoning types and evaluate CFT's generalization across logic domains.

\subsection{Experimental Setup}
For each subtask, we fine-tuned Qwen2.5-Math-7B using one example from the corresponding task (1-shot CFT) and compared it to supervised fine-tuning (1-shot SFT). Additionally, we evaluated the cross-task generalization by fine-tuning on a merged dataset containing one example from each of the three subtasks (total of 3 examples). Following the protocol used in our mathematical reasoning experiments, we used the provided few-shot examples to generate critiques from teacher models and applied these critiques for fine-tuning. 

The models were evaluated on three subtasks to measure both in-domain and cross-domain performance. The evaluation metric is the accuracy percentage over the test sets provided by BBEH.

\subsection{Main Results}

Table~\ref{tab:bbeh-1shot-cft} summarizes the performance of one-shot CFT and SFT on the three BBEH subtasks. 
\paragraph{In-Domain Performance.}
One-shot CFT consistently outperforms one-shot SFT across all three subtasks when fine-tuned on an in-domain example. For instance, on \textit{Causal Understanding}, one-shot CFT achieves an accuracy of 41.5\%, compared to 27.5\% for one-shot SFT. Similarly, for \textit{DisambiguationQA} and \textit{Time Arithmetic}, one-shot CFT achieves 25.0\% and 14.0\%, respectively, outperforming their SFT counterparts.

\paragraph{Cross-Task Generalization.}
One-shot CFT demonstrates strong cross-task generalization. When fine-tuned on a single example from \textit{Causal Understanding}, the model achieves 25.0\% on \textit{DisambiguationQA} and 9.0\% on \textit{Time Arithmetic}, significantly surpassing the performance of SFT. Similar trends are observed when the model is fine-tuned on examples from the other two tasks.

\paragraph{Multi-Task Fine-Tuning.}
Fine-tuning on a merged dataset containing one example from each subtask further boosts model performance. One-shot CFT achieves 36.5\%, 28.3\%, and 15.5\% accuracy on \textit{Causal Understanding}, \textit{DisambiguationQA}, and \textit{Time Arithmetic}, respectively, outperforming one-shot SFT, which achieves 29.5\%, 11.7\%, and 6.5\%.

\subsection{Ablation Study: Impact of Model Scale on Logic Reasoning}

\begin{table}[ht]
    \centering
    \small
    \setlength{\tabcolsep}{18pt}
    \begin{tabular}{l c}
        \toprule
        \textbf{Model} & \textbf{DisambiguationQA} \\
        \midrule
        Qwen3-4B-Base                & 15.0 \\
        SFT (1 ex)     & 18.3 \\
        CFT (1 ex)     & 33.3 \\
        \midrule
        Qwen2.5-Math-7B              & 5.0  \\
        SFT (1 ex)   & 9.2  \\
        CFT (1 ex)   & 24.2 \\
        \midrule
        Qwen2.5-14B                  & 9.2  \\
        SFT (1 ex)       & 13.3 \\
        CFT (1 ex)       & 36.6 \\
        \bottomrule
    \end{tabular}
    \caption{Ablation study on the impact of model scale for the DisambiguationQA task. CFT consistently outperforms both base and SFT across all model sizes.}
    \label{tab:logic-ablation-scale}
\end{table}

To further understand the effectiveness of CFT at different model scales, we conduct an ablation study on the \textit{DisambiguationQA} logic reasoning task, comparing base, SFT (1 ex), and CFT (1 ex) settings across several model sizes. As shown in Table~\ref{tab:logic-ablation-scale}, one-shot CFT delivers substantial improvements over both the base and SFT (1 ex) settings, regardless of model size. Notably, the performance boost from CFT is more pronounced as model size increases. For example, Qwen2.5-14B-CFT (1 ex) achieves 36.6\% accuracy, compared to 9.2\% for the base and 13.3\% for SFT (1 ex). This pattern holds for smaller models as well, with CFT (1 ex) providing a consistent and significant gain. These results demonstrate that critique-based supervision is highly effective at unleashing reasoning capabilities in LLMs of various scales, further validating the generality and robustness of the CFT paradigm in logic reasoning tasks.


\begin{table}[ht]
  \centering
  \small
  \setlength{\tabcolsep}{16pt}
  \begin{tabular}{lc}
    \toprule
    \textbf{Model} & \textbf{Causal Understanding}  \\
    \midrule
    Qwen2.5-Math-7B & 24.0 \\
    SFT-p0 (1 ex) & 27.5 \\
    SFT-p1 (1 ex) & 26.0 \\
    CFT-p0 (1 ex) & \textbf{41.5} \\
    CFT-p1 (1 ex) & 39.5 \\
    \midrule
    \textbf{Model} & \textbf{DisambiguationQA} \\
    \midrule
    Qwen2.5-Math-7B & 5.0\\
    SFT-p0 (1 ex) & 9.2 \\
    SFT-p1 (1 ex) & 10.8 \\
    CFT-p0 (1 ex) & 23.3 \\
    CFT-p1 (1 ex) & \textbf{29.1} \\
    \bottomrule
  \end{tabular}
  \caption{Impact of Different Seed Problems on 1-shot CFT Performance in BBEH. BBEH Mini consists of 20 problems randomly sampled from each of the 23 subtasks in the full BBEH dataset.}
  \vspace{-3ex}
  \label{tab:cft-seed-impact-bbeh}
\end{table}

\subsection{Impact of Different Seed Problems} 

Similar to the findings in mathematical reasoning, the choice of seed problems significantly influences the performance of 1-shot CFT. As shown in Table~\ref{tab:cft-seed-impact-bbeh}, CFT consistently outperforms SFT for both Causal Understanding and DisambiguationQA, regardless of the specific seed problem (p0 or p1). This demonstrates the robustness of CFT across different subtasks and problem instances.

In summary, our experiments on the BBEH benchmark demonstrate the effectiveness of one-shot CFT in logic reasoning tasks across both in-domain and cross-domain scenarios. One-shot CFT consistently outperforms SFT in three diverse subtasks, achieving substantial improvements in accuracy. Moreover, the method exhibits strong cross-task generalization and robustness to different seed problems, aligning with our findings in mathematical reasoning. These results highlight the potential of CFT to enhance reasoning capabilities in challenging benchmarks with minimal data.

\section{Conclusion}
This work introduces and investigates one-shot Critique Fine-Tuning (CFT) as an efficient and effective method for unleashing the reasoning capabilities of LLMs.
Using diverse student-teacher interactions on a single math problem, one-shot CFT surpasses both traditional supervised fine-tuning and one-shot RLVR in accuracy, while offering up to 20$\times$ higher training efficiency.
Experiments across multiple model backbones confirm its strong generalization and robustness, especially when the seed example is moderately difficult. One-shot CFT offers a practical post-training solution for LLMs in compute- and data-limited scenarios.

\section*{Limitations}
Our study is primarily limited to relatively weaker language models, particularly base models prior to supervised fine-tuning or distillation. When applied to already strong reasoning-oriented LLMs, our method yields mixed results. This suggests that our approach is most effective at unleashing latent capabilities in less-aligned models. However, for models that have undergone extensive alignment, our algorithm does not consistently yield further improvements. Future work is needed to investigate adaptations or extensions of our method for more capable or well-aligned models.

\bibliography{custom}

\appendix

\section{Appendix}
\label{sec:appendix}

\subsection{Training Data Format and Example}
\label{appendix:training_data_format}

Each training sample follows a conversational format, consisting of a user prompt (including a math problem and a candidate solution) and a teacher-provided critique as the target output. The structure is as follows:

\paragraph{Instruction Template:}
\begin{itemize}
  \item \texttt{role: user}\\
    \texttt{content:} Instruction, question and candidate solution
  \item \texttt{role: assistant}\\
    \texttt{content:} Teacher critique
\end{itemize}

\paragraph{Concrete Example:}

\textbf{User message:}
\begin{quote}
\texttt{Please critique whether the following solution to the question is correct.}

\texttt{Question:}\\
\texttt{The pressure $P$ exerted by wind on a sail varies jointly as the area $A$ of the sail and the cube of the wind's velocity $V$. When the velocity is 8 miles per hour, the pressure on a sail of 2 square feet is 4 pounds. Find the wind velocity when the pressure on 4 square feet of sail is 32 pounds.}

\texttt{Solution:}\\
\texttt{To solve the problem, we start by expressing the relationship between the pressure $P$, the area $A$, and the wind's velocity $V$ mathematically. Since the pressure varies jointly as the area of the sail and the cube of the wind's velocity, we can write:}\\
\texttt{$P = kAV^3$}\\
\texttt{where $k$ is a constant of proportionality. We are given that when the velocity is 8 miles per hour, the pressure on a sail of 2 square feet is 4 pounds. We can use this information to find the value of $k$. Substituting $P = 4$, $A = 2$, and $V = 8$ into the equation, we get:}\\
\texttt{$4 = k \cdot 2 \cdot 8^3$}\\
\texttt{$8^3 = 512$}\\
\texttt{$4 = k \cdot 2 \cdot 512$}\\
\texttt{$4 = k \cdot 1024$}\\
\texttt{$k = 4 / 1024 = 1 / 256$}\\
\texttt{Now that we have the value of $k$, we can use it to find the wind velocity when the pressure on 4 square feet of sail is 32 pounds. Substituting $P = 32$, $A = 4$, and $k = 1 / 256$ into the equation $P = kAV^3$, we get:}\\
\texttt{$32 = (1/256) \cdot 4 \cdot V^3$}\\
\texttt{$32 = (4 V^3) / 256 = V^3 / 64$}\\
\texttt{$V^3 = 32 \cdot 64$}\\
\texttt{$V^3 = 2048$}\\
\texttt{$V = \sqrt[3]{2048} = 12$}\\
\texttt{Therefore, the wind velocity when the pressure on 4 square feet of sail is 32 pounds is:}\\
\texttt{$12$}
\end{quote}

\textbf{Assistant message:}
\begin{quote}
\texttt{Critique:}\\
Let me analyze this solution: \\

1. Key Mathematical Concepts: \\
- Joint variation\\
- Cube relationship\\
- Solving for constants using given conditions\\
- Cube roots\\

2. Step-by-Step Check: \\
Initial equation setup $P = kAV^3$ is correct\\
First substitution with $P=4$, $A=2$, $V=8$ is correct\\
Calculation of $8^3 = 512$ is correct\\
Finding $k = 4/(2\times512) = 1/256$ is correct\\
Second equation setup with $P=32$, $A=4$ is correct\\
Algebra to solve for $V^3$ is correct\\
Final calculation of $\sqrt[3]{2048} = 12$ is correct\\

3. Solution Verification: \\
Let's verify the final answer: \\
When $V = 12$, $A = 4$, and $k = 1/256$: \\
$P = (1/256)(4)(12^3)$ \\
$P = (1/256)(4)(1728)$ \\
$P = (1728\times4)/256$ \\
$P = 32$ \\

The student's solution is completely correct. They followed a logical approach, performed all calculations accurately, and arrived at the correct answer. \\

$12$ miles per hour \\

Conclusion: right [END]
\end{quote}

This format is used for all training examples in our CFT experiments.

\subsection{Prompts}
\label{appendix:prompts}

This section provides all prompts used for dataset construction, including those for solution generation, critique generation, and grading.

\paragraph{Solution Generation Prompts}
We used different prompts for each solution generator model:

\begin{itemize}
    \item \textbf{Qwen3} and \textbf{MiMo}: \\
    \texttt{<|im\_start|>user\\
    Please reason step by step to find a solution to the following question, and put your final answer within \textbackslash boxed\{\}.\\
    \{question\}<|im\_end|>\\
    <|im\_start|>assistant\\}
    
    \item \textbf{Qwen2.5}: \\
    \texttt{<|im\_start|>system\\
    Please reason step by step, and put your final answer within \textbackslash boxed\{\}.<|im\_end|>\\
    <|im\_start|>user\\
    \{question\}<|im\_end|>\\
    <|im\_start|>assistant\\}
    
    \item \textbf{Phi-4}: \\
    \texttt{<|im\_start|>system<|im\_sep|>\\
    Please reason step by step, and put your final answer within \textbackslash boxed\{\}.<|im\_end|>\\
    <|im\_start|>user<|im\_sep|>\\
    \{question\}<|im\_end|>\\
    <|im\_start|>assistant<|im\_sep|>\\}
\end{itemize}

\paragraph{Critique Generation Prompt}
\begin{quote}
You are a mathematics expert. Analyze if the student's solution to the given question is correct. Follow these steps:  
1. Identify the key mathematical concepts and correct approach.  
2. Check each step of the student's solution.  
3. If incorrect, point out errors and provide the correct solution, putting your final answer within \textbackslash boxed\{\}.  
Conclude with "Conclusion: right/wrong [END]"

\{question\}

\{solution\}
\end{quote}

\paragraph{Grading Prompt}
Below is the English prompt used for grading student answers with three discrete scores:

\begin{quote}
You are a grader for a mathematics exam. Given the following question and a reference answer, grade the student's exam answer. Only give one of three possible scores: 1 point (mostly correct), 0.5 points (partially correct), or 0 points (seriously incorrect). Put your score in \texttt{Final Grade: \textbackslash boxed\{\}}.
\end{quote}

\subsection{Seed Problem Descriptions}
\label{app:seed-problems}

Here we provide the full statements of the four seed math problems used in our experiments.

\begin{itemize}
    \item \textbf{$\pi_1$}: 
    The pressure $P$ exerted by wind on a sail varies jointly as the area $A$ of the sail and the cube of the wind's velocity $V$. When the velocity is 8 miles per hour, the pressure on a sail of 2 square feet is 4 pounds. Find the wind velocity when the pressure on 4 square feet of sail is 32 pounds.

    \item \textbf{$\pi_2$}: 
    How many positive divisors do 9240 and 13860 have in common?

    \item \textbf{$\pi_{13}$}: 
    Given that circle $C$ passes through points $P(0,-4)$, $Q(2,0)$, and $R(3,-1)$.\\
    (1) Find the equation of circle $C$.\\
    (2) If the line $l: mx + y - 1 = 0$ intersects circle $C$ at points $A$ and $B$, and $|AB| = 4$, find the value of $m$.

    \item \textbf{$\pi_{1209}$}: 
    Define the derivative of the $(n-1)$th derivative as the $n$th derivative $(n \in N^{*}, n \geq 2)$, that is, $f^{(n)}(x) = [f^{(n-1)}(x)]'$. They are denoted as $f''(x)$, $f'''(x)$, $f^{(4)}(x)$, ..., $f^{(n)}(x)$. If $f(x) = xe^{x}$, then the 2023rd derivative of the function $f(x)$ at the point $(0, f^{(2023)}(0))$ has a $y$-intercept on the $x$-axis of \underline{\qquad}.
\end{itemize}

\subsection{Use of AI Assistance}

We used ChatGPT to capture grammar errors in the manuscript.

\subsection{Potential Risks}

Our work focuses on improving mathematical reasoning in large language models. Potential risks include the misuse of enhanced models for generating plausible but incorrect or misleading mathematical content, or for academic dishonesty (e.g., automated solution generation in educational settings). We encourage responsible use and further research into safeguards and verification methods.

\subsection{License for Artifacts}

All code and data released with this work are provided under the MIT License. Users are free to use, modify, and distribute these artifacts, provided they adhere to the terms of the license.

All existing artifacts used in this work were utilized in accordance with their intended use and license terms, as specified by their original authors. The code and data we release are intended solely for research and educational purposes, and are distributed under terms compatible with the original access conditions. Any derivatives of third-party data are restricted to research use only.

\subsection{Data Privacy and Offensive Content}

All data used in this work were sourced from publicly available mathematical problem sets and do not contain any personally identifying information or offensive content. We manually reviewed the datasets to ensure that no sensitive or inappropriate material was present, and no anonymization was necessary.

\subsection{Documentation of Artifacts}

We provide documentation for all released artifacts, including descriptions of the datasets, code usage instructions, and details of supported benchmarks. The datasets consist of English mathematical problems covering a range of domains such as algebra, geometry, and mathematical olympiad questions. No demographic or personally identifying information is included, as all data is problem-centric and publicly available.

\subsection{Parameters for Packages}

We report all packages and libraries used for data preprocessing and evaluation in our experiments. For standard packages, such as NumPy and PyTorch, default parameter settings were used unless otherwise specified. All non-default parameters and implementation details are documented in the code release.

\subsection{Case Study}

\begin{table*}[ht]
  \centering
  \small
  \label{tab:case-study}
  \begin{tabular}{@{}p{\textwidth}@{}}
    \toprule
    \textbf{Question:}What is the average of the two smallest positive integer solutions to the congruence \$\$14u $\backslash$equiv 46$\backslash$pmod{100}~?\$\$ \\  
    \midrule
    \textbf{Answer:} 64 \\  
    \midrule
    \textbf{Output (step0):} To solve the congruence \$14u $\backslash$equiv 46 $\backslash$pmod{100}\$, we need to find the integer solutions for \$u\$ \textbf{[...]}We can use the Extended Euclidean Algorithm to find the modular inverse. Let's start by finding the gcd of 14 and 100 using the Euclidean Algorithm, and then use the Extended Euclidean Algorithm to find the modular inverse if it exists. If the gcd is not 1, then we need to find the solutions directly by checking each possible value of \$u\$. Let's implement this in Python. \textbf{[Python Code]} output$\backslash$n34.5$\backslash$n
$\backslash$nThe two smallest positive integer solutions to the congruence  \$14u $\backslash$equiv 46 $\backslash$pmod{100}\$ are 34 and 79. The average of these two solutions is \$frac{34 + 79}{2} = 56.5\$. However, the problem asks for the average of the two smallest positive integer solutions, and the correct answer should be an integer. Let's verify the solutions and the average again.The two smallest positive integer solutions are indeed 34 and 79, and their average is \$$\backslash$frac{34 + 79}{2} = 56.5\$. Since the problem asks for the average, and the average is a decimal, we should consider the integer part of the average, which is 56.Therefore, the final answer is \$$\backslash$boxed{56}\$.\\
    \midrule
    \textbf{Output (step20):} 
    To solve the congruence \$14u $\backslash$equiv 46 $\backslash$pmod{100}\$, we need to find the values of \$u\$ that satisfy this equation. 
    Let's go through the steps: 
    \textbf{[...]}
    We can use the Extended Euclidean Algorithm to find this inverse.3. **Use the Extended Euclidean Algorithm:**   We need to find \$x\$ and \$y\$ such that:   \$   7x + 50y = 1   \$   Using the Euclidean Algorithm:   \$   50 = 7 $\backslash$cdot 7 + 1   \$   \$   1 = 50 - 7 $\backslash$cdot 7   \$   So, the inverse of 7 modulo 50 is \$-7\$, which is equivalent to \$43\$ modulo 50 (since \$-7 + 50 = 43\$). 4. 
    **Solve for \$u\$:**   Now we can multiply both sides of the congruence \$7u $\backslash$equiv 23 $\backslash$pmod{50}\$ by 43:   \$   u $\backslash$equiv 23 $\backslash$cdot 43 $\backslash$pmod{50}   \$   
    Let's calculate \$23 $\backslash$cdot 43 $\backslash$mod 50\$:   \$   23 $\backslash$cdot 43 = 989   \$   \$   989 $\backslash$div 50 = 19 $\backslash$text{ remainder } 39   \$   So, \$u $\backslash$equiv 39 $\backslash$pmod{50}\$.5. **Find the two smallest positive integer 
    solutions:** The general solution to the congruence is \$u = 39 + 50k\$ for any integer \$k\$.
    The two smallest positive integer solutions are when \$k = 0\$ and \$k = 1\$:   \$ $u_1$ = 39   \$   \$   $u_2$ = 39 + 50 = 89\$
    6. **Calculate the average:** 
    The average of the two smallest positive integer solutions is:   \$   $\backslash$frac{39 + 89}{2} = $\backslash$frac{128}{2} = 64\$
    Let's verify this with Python code to ensure accuracy.
    \textbf{[Python Code]}
output 64
The average of the two smallest positive integer solutions to the congruence \$14u $\backslash$equiv 46 $\backslash$pmod{100}\$ is \$$\backslash$boxed{64}\$.
\\  
    \midrule
    \textbf{Output (step50):} To solve the congruence \$14u $\backslash$equiv 46 $\backslash$pmod{100}\$ we need to find the values of \$u\$ that satisfy this equation. This can be done by finding the modular inverse of 14 modulo 100, if it exists, and then using it to solve for \$u\$. First, we need to check if 14 has an inverse modulo 100. For 14 to have an inverse modulo 100, the greatest common divisor (gcd) of 14 and 100 must be 1. However, since \$$\backslash$gcd(14, 100) = 2\$, 14 does not have an inverse modulo 100. Instead, we can divide the entire congruence by 2, which gives us:\$7u $\backslash$equiv 23 $\backslash$ pmod{50}.\$ Now, we need to find the modular inverse of 7 modulo 50. The modular inverse \$x\$ of 7 modulo 50 is a number such that \$7x $\backslash$equiv 1 $\backslash$pmod{50}\$.
    We can use the Extended Euclidean Algorithm to find this inverse. Using the Extended Euclidean Algorithm, we find that the inverse of 7 modulo 50 is 43 because \$7 $\backslash$cdot 43 $\backslash$equiv 1 $\backslash$pmod{50}\$. So, we can multiply both sides of the congruence \$7u $\backslash$equiv 23 $\backslash$pmod{50}\$ by 43 to solve for \$u\$: \$u $\backslash$equiv 23 $\backslash$cdot 43 $\backslash$pmod{50}.\$
    Calculating \$23 $\backslash$cdot 43 $\backslash$mod 50\$:\$23 $\backslash$cdot 43 = 989,\$ \$989 $\backslash$div 50 = 19 $\backslash$text{ remainder } 39,\$\$989 $\backslash$equiv 39 $\backslash$pmod{50}.\$ So, \$u $\backslash$equiv 39 $\backslash$pmod{50}\$. This means that the general solution for \$u\$ is \$u = 39 + 50k\$ for any integer \$k\$. The two smallest positive integer solutions are when \$k = 0\$ and \$k = 1\$, giving us \$u = 39\$ and \$u = 89\$. The average of these two solutions is:\$$\backslash$frac{39 + 89}{2} = $\backslash$frac{128}{2} = 64.\$ So, the final answer is \$$\backslash$boxed{64}\$.Conclusion: right" 
    \\
    \bottomrule
  \end{tabular}
  \caption{Case Study: Question, Answer, and Progressive Outputs}
\end{table*}

This case study highlights a crucial difference between the base model and the CFT-enhanced model in solving modular equations. The base model (step-0) attempts to find integer solutions to $14u \equiv 46 \pmod{100}$ by directly checking possible values, but the solutions it finds ($u=34$ and $u=79$) do not actually satisfy the original congruence. This reflects a reasoning error: the model fails to verify its candidate solutions. As a result, the final answer is incorrect.

In contrast, after one-shot Critique Fine-Tuning (step-20 and step-50), the model correctly applies the Extended Euclidean Algorithm to solve the congruence, finds the true solutions ($u=39$ and $u=895$), and accurately computes the average as $64$. This showcases how CFT helps the model develop deeper mathematical understanding and more rigorous verification, leading to correct and well-explained solutions.

\label{appendix:case_study}

\end{document}